\documentclass[conference]{IEEEtran}

\usepackage{fancyhdr}
\usepackage{setspace}
\usepackage[normalem]{ulem}
\usepackage[hyphens]{url}
\usepackage[caption=false]{subfig}
\usepackage{amsmath}
\usepackage{booktabs}
\usepackage{multirow}
\usepackage{threeparttable}
\usepackage{array,multirow}
\usepackage{graphicx}
\usepackage{makecell}
\usepackage{epsfig}
\usepackage{enumitem}
\usepackage{cleveref}
\usepackage[square,sort&compress,comma,numbers]{natbib}

\usepackage{footnote}
\makesavenoteenv{tabular}
\makesavenoteenv{table}

\title
{
    Towards Closing the Energy Gap Between HOG and CNN Features for Embedded Vision
}

\author
{
    \IEEEauthorblockN   {   Amr Suleiman*, Yu-Hsin Chen*, Joel Emer, Vivienne Sze}
    \IEEEauthorblockA   {   Department of Electrical Engineering and Computer Science, Massachusetts Institute of Technology\\ 
                            \{suleiman, yhchen, jsemer, sze\}@mit.edu\\
                \textit{*These authors contributed equally to this work}}
}

\IEEEspecialpapernotice{(Invited Paper)}

\begin{document}
\maketitle


\begin{abstract}


Computer vision enables a wide range of applications in robotics/drones, self-driving cars, smart Internet of Things, and portable/wearable electronics. For many of these applications, local embedded processing is preferred due to privacy and/or latency concerns. Accordingly, energy-efficient embedded vision hardware delivering real-time and robust performance is crucial. While deep learning is gaining popularity in several computer vision algorithms, a significant energy consumption difference exists compared to traditional hand-crafted approaches. In this paper, we provide an in-depth analysis of the computation, energy and accuracy trade-offs between learned features such as deep Convolutional Neural Networks (CNN) and hand-crafted features such as Histogram of Oriented Gradients (HOG). This analysis is supported by measurements from two chips that implement these algorithms. Our goal is to understand the source of the energy discrepancy between the two approaches and to provide insight about the potential areas where CNNs can be improved and eventually approach the energy-efficiency of HOG while maintaining its outstanding performance accuracy.

\end{abstract}


\section{Introduction}
\label{sec:introduction}

Computer vision (CV) is a critical technology to numerous smart embedded systems, such as advanced driver assistant systems, autonomous cars/drones, and robotics. It extracts meaningful information from visual data for further decision making. However, many modern CV algorithms require high computational complexity, which makes their deployment on battery-powered devices challenging due to the tight energy constraints. Near-sensor visual data processors should consume under 1nJ/pixel with a logic gate counts of around 1000 kgates and a memory capacity of few hundred kBytes in order to be comparable with video codecs, which are present in most cameras~\cite{7063063}. For many applications, offloading computation to the cloud is also undesirable because of latency, connectivity, and security limitations. Thus, dedicated energy-efficient CV hardware becomes very crucial.

\begin{figure}
    \begin{center}
        \includegraphics[width=0.9\linewidth]{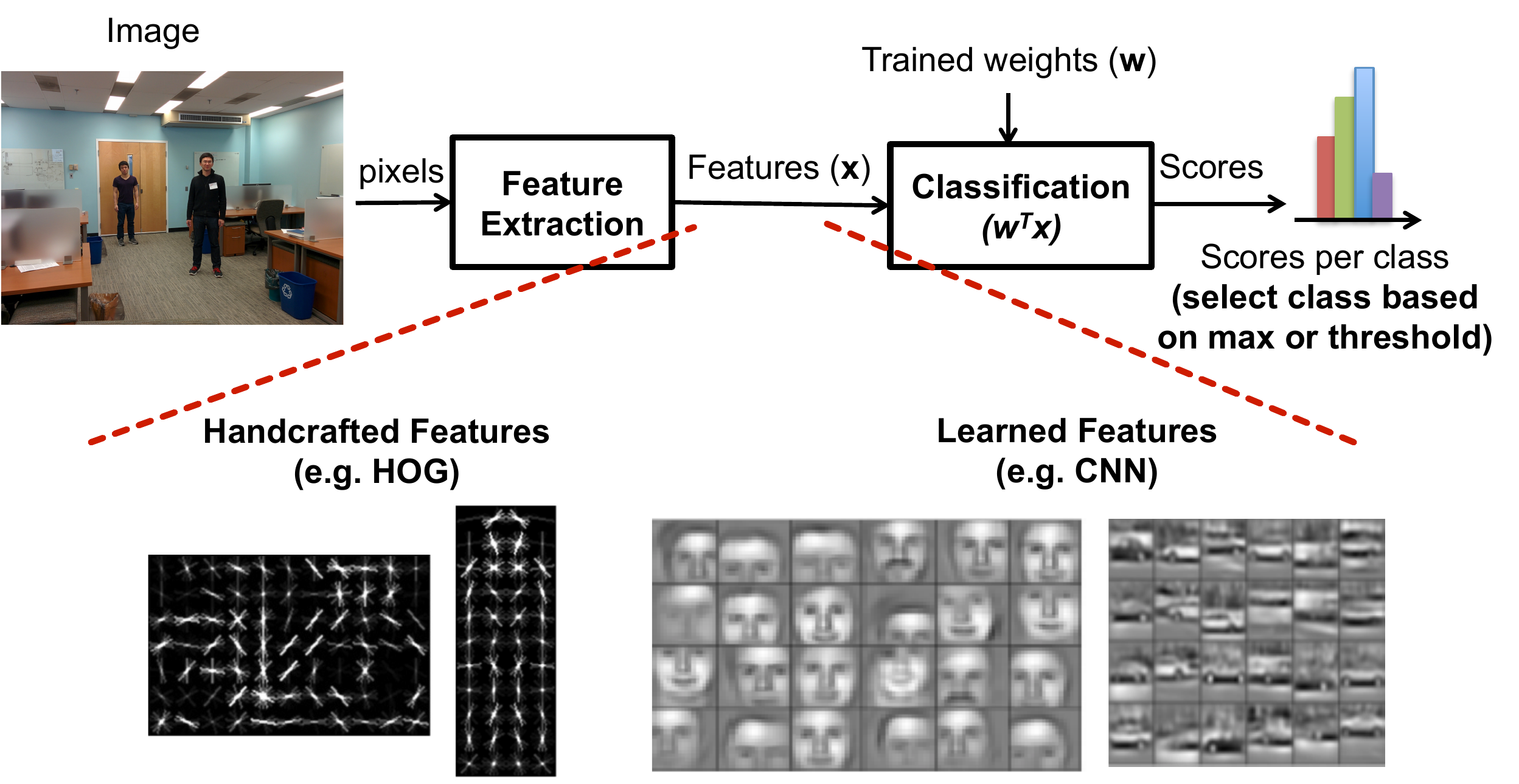}
        \caption{General processing pipeline for object classification and detection.
                }
        \vspace{-10pt}                
        \label{fig:cv_processing_pipeline}
    \end{center}
\end{figure}

Feature extraction is the first processing step in most CV tasks, such as object classification and detection (Fig.~\ref{fig:cv_processing_pipeline}). It transforms the raw pixels into a high-dimensional space, where only meaningful and distinctive information is kept. Traditionally, features are designed by experts in the field through a hand-crafted process. For instance, many well-known \emph{hand-crafted features} use image gradients, such as histogram of oriented gradients (HOG)~\cite{dalal2005histograms} and scale invariant feature transform (SIFT)~\cite{lowe1999object}, based on the fact that human eyes are sensitive to edges. In contrast, \emph{learned features} learn a representation with the desired characteristics directly from data using deep convolutional neural networks (CNNs)~\cite{iscas2010-lecun}. Learned features are gaining popularity, as they are outperforming hand-crafted features in many CV tasks~\cite{nips2012-krizhevsky, cvpr2014-girshick}.

The differences in design between hand-crafted and learned features lead to not only different performance in applications, but also different hardware implementation considerations, which have a strong implication for energy efficiency. In general, hardware implementations for hand-crafted features are widely understood to be more energy-efficient than learned features. However, there is no analysis that explains the energy gap between the two types of features. Also, an open question is whether the energy gap can be closed in the future.

In this paper, we will provide an in-depth analysis on the causes for the energy gap between hand-crafted and learned features. We use results from two actual chip designs:~\cite{suleiman201658} implements the hand-crafted feature using HOG, and~\cite{isscc2016-chen} implements the learned feature using CNN. Both chips use 65nm CMOS technology and have similar hardware resource utilization in terms of logic gate count and memory capacity. Based on the insights derived from the two implementations, we will discuss techniques to help close the energy gap.


\section{Feature Extraction Hardware Implementations}
\label{sec:hardware_implementation}

\subsection{Hardware for Hand-crafted Feature: HOG}

\begin{figure}
    \begin{center}
        \includegraphics[width=0.85\linewidth]{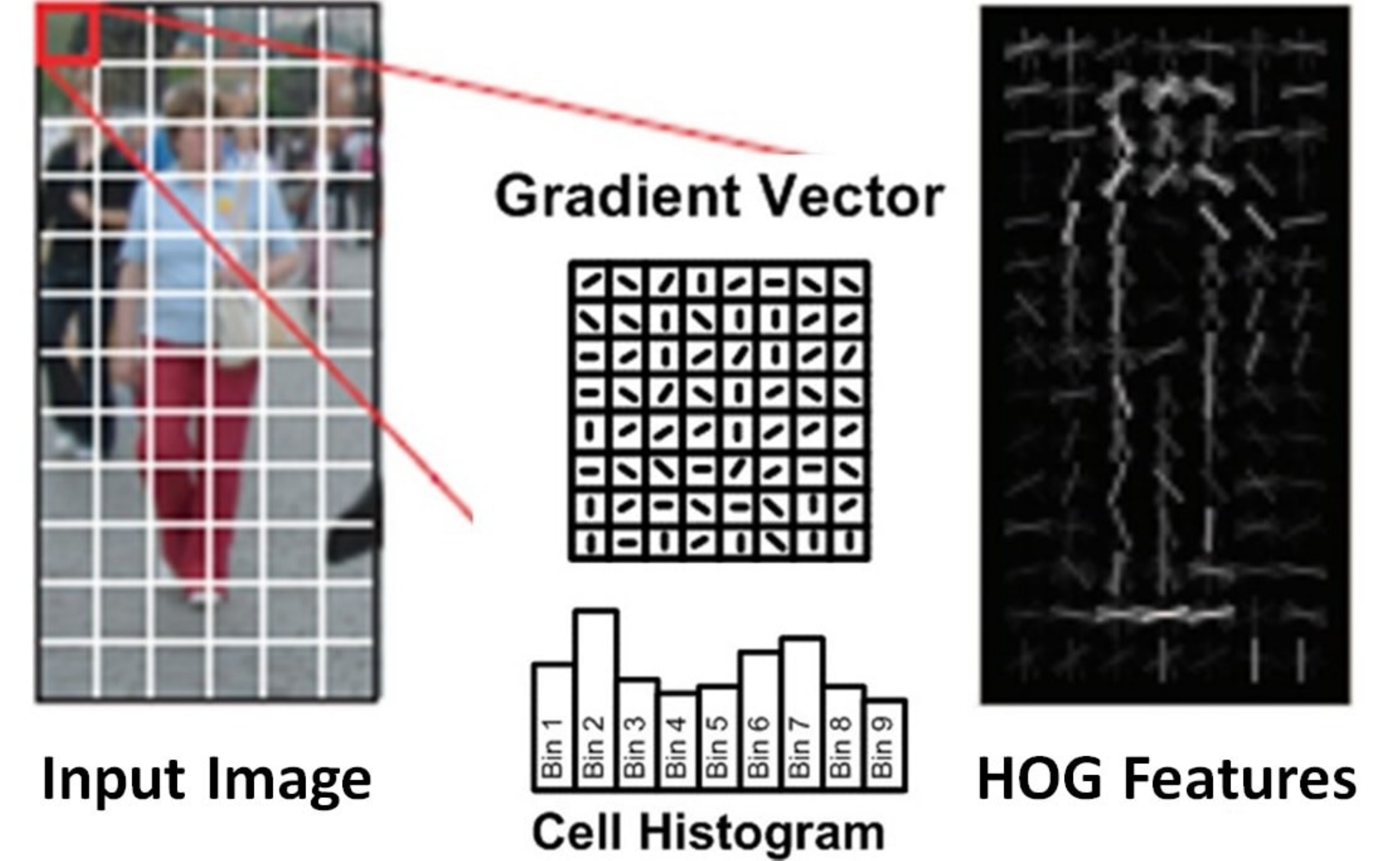}
        \caption{Feature extraction with histogram of oriented gradients (HOG).
                }
        \vspace{-10pt}                
        \label{fig:hog_simple}
    \end{center}
\end{figure}

The chip presented in~\cite{suleiman201658} implements the entire object detection pipeline based on deformable parts models (DPM)~\cite{girshick2012discriminatively} for high throughput and low power embedded vision applications. DPM extracts HOG features from the input image, and localizes objects by sweeping the features with support vector machine (SVM) classifiers~\cite{cristianini2000introduction}. Fig.~\ref{fig:hog_simple} shows the features extraction process using HOG: the image is divided into non-overlapping 8$\times$8 pixel cells. A 9-bin histogram is then generated for each cell using gradients orientations. The histograms are further normalized with respect to the surrounding cells for robustness against illumination and texture variations. Since HOG features are not scale invariant, the features are extracted over an image pyramid to achieve multi-scale detection.

\subsection{Hardware for Learned Feature: CNN}

\begin{figure}
    \begin{center}
        \includegraphics[width=0.9\linewidth]{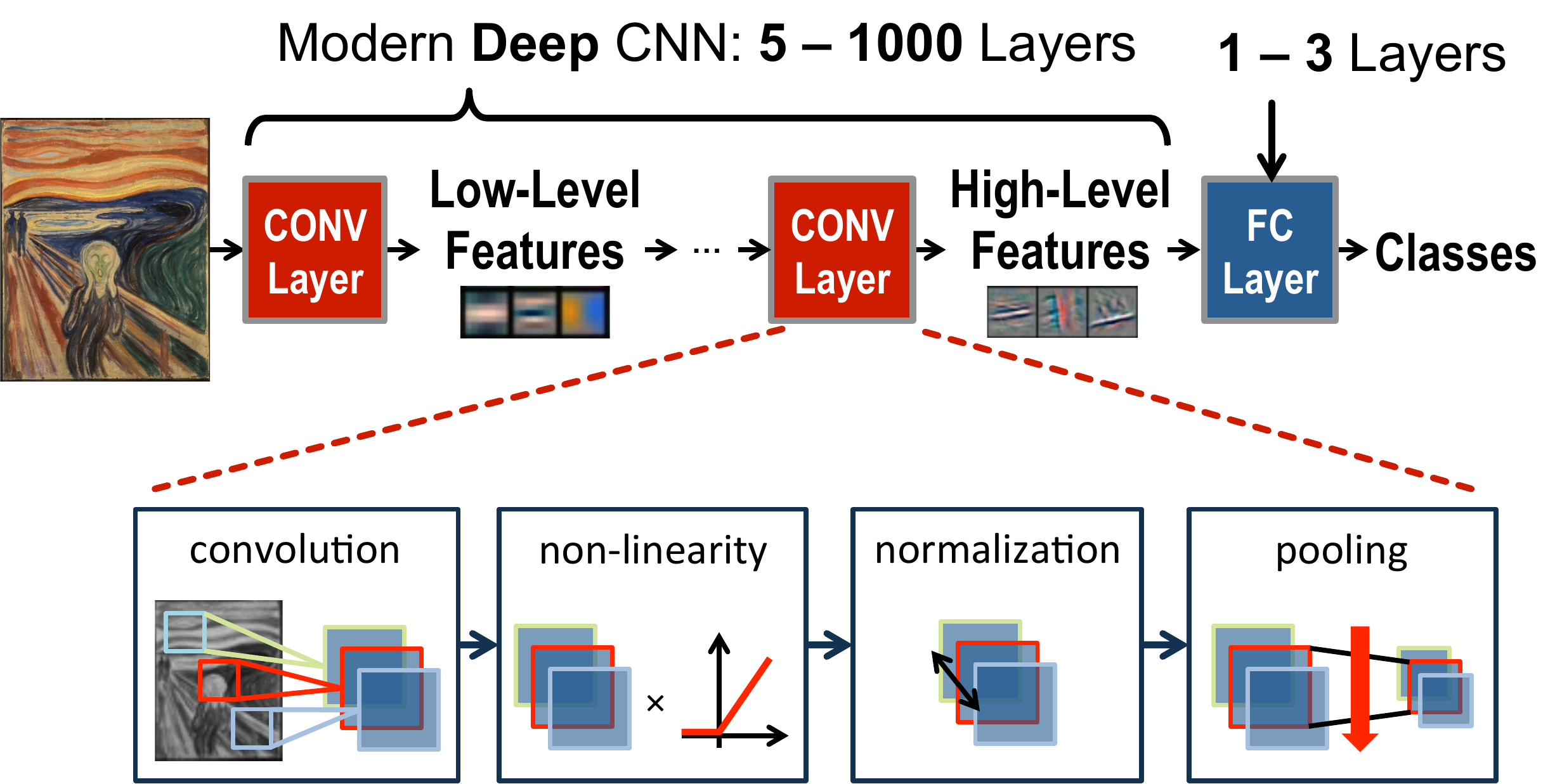}
        \caption{General processing pipeline of CNN.
                }
        \vspace{-10pt}                
        \label{fig:DNN}
    \end{center}
\end{figure}

The second chip presented in~\cite{isscc2016-chen}, called Eyeriss, is an energy-efficient accelerator for deep CNNs. Fig.~\ref{fig:DNN} shows a general CNN processing pipeline, consisting mainly of a series of convolutional (CONV) layers. In each CONV layer, a collection of 3D filters are applied to the input images or feature maps to generate output feature maps, which are then used as the input to the next CONV layer. Eyeriss is programmable in terms of the size and shape of filters as well as number of layers. Therefore, it can accommodate different CNN architectures, such as AlexNet~\cite{nips2012-krizhevsky} and VGG-16~\cite{iclr2015-simonyan}.

\subsection{Performance Comparison}

Table~\ref{tab:compare_chips} shows the hardware specification and measured performance of the two designs for feature extraction. For CNN, Eyeriss is programmed to run two CNN models (five CONV layers of AlexNet and thirteen CONV layers of VGG-16) to demonstrate the hardware performance variations of running different CNN features. Both chip designs use around 1000 kgates and 150 kB memory. While Eyeriss achieves approximately the same computation throughput (i.e., GOPS) as the HOG design when running AlexNet features, HOG processes 35$\times$ more input pixels per second; the gap is even larger between HOG and VGG-16 features. This is due to the differences in computational complexity (i.e., operations per pixel) between different features as shown in Table~\ref{tab:feat_comp}. However, the HOG design also consumes around 10$\times$ less power than Eyeriss. Thus, the HOG hardware consumes 311$\times$ and 13,486$\times$ less energy per pixel than Eyeriss running AlexNet and VGG-16 features, respectively. In terms of energy per operation, the HOG hardware is 10$\times$ less than Eyeriss. In Section~\ref{sec:hardware_considerations}, we will discuss the cause of this energy gap.

\begin{table}
    \begin{center}
        \scriptsize
        \caption{Hardware specification and measured performance of hand-crafted feature HOG~\cite{suleiman201658} and learned feature CNN~\cite{isscc2016-chen}.}
        \begin{tabular}{|l|c|c|c|} 
            \hline
                                            & \cite{suleiman201658}                 & \multicolumn{2}{c|}{\cite{isscc2016-chen}}             \\
            \hline
            \multirow{2}{*}{\textbf{Implemented Feature}}    & \multirow{2}{*}{HOG} & CNN               & CNN \\ 
                                            &                                       & (AlexNet)         & (VGG-16) \\
            \hline
            \textbf{Technology}             & 65nm                                  & \multicolumn{2}{c|}{65nm} \\ 
            \hline
            \textbf{Gate Counts (kgates)}   & 893.0                                 & \multicolumn{2}{c|}{1176.0} \\ 
            \hline
            \textbf{Memory (kB)}            & 159.0                                 & \multicolumn{2}{c|}{181.5} \\ 
            \hline
            \textbf{Multiplier Bitwidth}    & 5$\times$11 -- 22$\times$22           & \multicolumn{2}{c|}{16$\times$16} \\
            \hline
            \textbf{Throughput (Mpixels/s)} & 62.5                                  & 1.8               & 0.04 \\ 
            \hline
            \textbf{Throughput (GOPS)}      & 46.0                                  & 46.2              & 21.4 \\ 
            \hline
            \textbf{Power (mW)}             & 29.3                                  & 278.0             & 236.0 \\ 
            \hline
            \textbf{DRAM access (B/pixel)}  & 1.0                                   & 74.7              & 2128.6 \\ 
            \hline
            \textbf{Energy Efficiency (nJ/pixel)}      & 0.5                        & 155.5             & 6742.9 \\ 
            \hline
            \textbf{Energy Efficiency (GOPS/W)} & 1570.0                            & 166.2             & 90.7 \\
            \hline
        \end{tabular}
        \label{tab:compare_chips}
    \end{center}
    \vspace{-10pt}
\end{table}

\begin{table}
    \begin{center}
        \small
        \caption{Computational complexity comparison between features}
        \begin{tabular}{|l|c|c|c|} 
            \hline
            \multicolumn{2}{|c|}{\textbf{Feature}}      & \textbf{GOP/Mpixel}   & \textbf{Ratio}\\ 
            \hline
            Hand-crafted                & HOG           & 0.7                   & 1.0$\times$   \\ 
            \hline
            \multirow{2}{*}{Learned}    & CNN (AlexNet) & 25.8                  & 36.9$\times$  \\ 
            \cline{2-4}
                                        & CNN (VGG-16)  & 610.3                 & 871.9$\times$ \\ 
            \hline
        \end{tabular}
        \label{tab:feat_comp}
    \end{center}
    \vspace{-10pt}
\end{table}

\subsection{Accuracy vs. Energy Efficiency}

While there is a large energy gap between the hardware for hand-crafted and learned features, the performance differences in applications, e.g., accuracy, between the features should also be taken into account. To measure accuracy, we use the features for object detection, which localizes and classifies objects in provided images (i.e., the outputs are the class and coordinates of each object in the images). The mean average precision (mAP) metric is used to quantify detection accuracy.

\begin{figure}
    \begin{center}
        \includegraphics[width=0.9\linewidth]{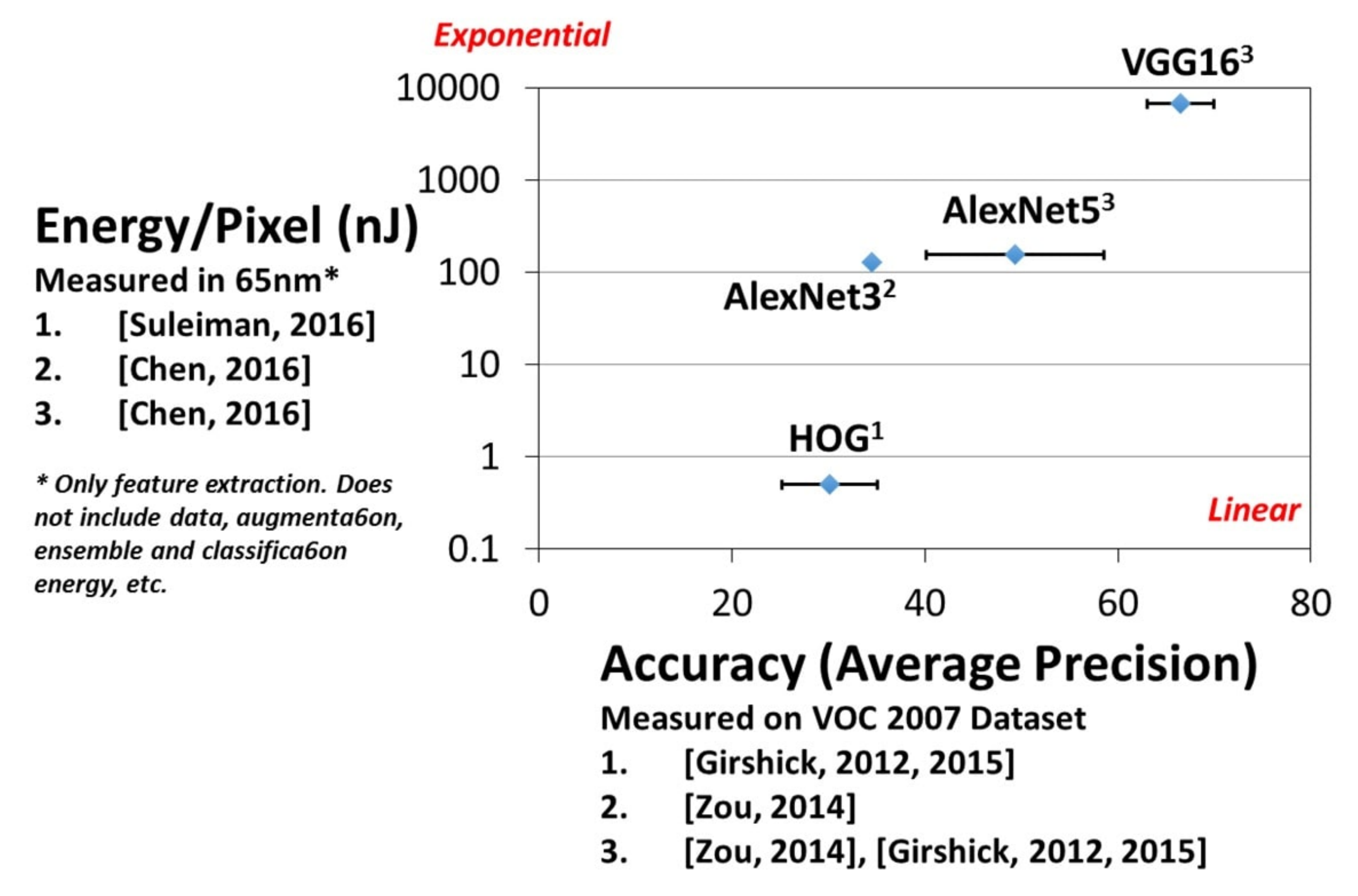}
        \caption{    Energy versus accuracy trade-off comparison for hand-crafted and learned features.
                }
        \vspace{-10pt}                
        \label{fig:hand_vs_dnn}
    \end{center}
\end{figure}

Fig.~\ref{fig:hand_vs_dnn} shows the trade-off between detection accuracy and energy consumption. Note that the vertical axis is logarithmic. All reported detection accuracy numbers are measured on PASCAL VOC 2007 dataset~\cite{pascal-voc-2007}, which is a widely used image dataset containing 20 different object classes in 9,963 images. In order to achieve the same detection accuracy of HOG features, it only requires the features extracted from the first three CONV layers of AlexNet instead of five~\cite{zou2014dnp}, but it comes at the cost of 100$\times$ higher energy consumption per pixel to generate the features. Fortunately, mAP can be nearly doubled by using the ouput of all five layers of AlexNet with minimal increase in energy consumption (only 22\%). Even higher mAP has been demonstrated by using VGG-16 features~\cite{girshick2015frcnn}. However, the accompanying energy consumption per pixel becomes four orders of magnitude higher than using HOG features. If we look at the minimum hardware energy consumption required for a given accuracy in Fig.~\ref{fig:hand_vs_dnn}, we roughly see that a linear increase in detection accuracy requires an exponential increase in energy consumption per pixel.


\section{Hardware Implementation Considerations}
\label{sec:hardware_considerations}


One of the key factors that differentiates the two types of features in hardware implementation is \emph{programmability}, which defines the flexibility of a feature to be configured to deal with different data statistics or different tasks. While hand-crafted features can achieve a certain degree of invariance to data variations (e.g., images with different exposures), they are mostly designed for very specific tasks with known data statistics, leaving little to be programmed when deployed. In contrast, learned features isolate algorithm design from learning the data statistics, which leaves room for programmability to take advantage of the flexibility. We categorize programmability into two types: Programmability of Hyper-Parameters (PoHP) and Programmability of Parameters (PoP).


\subsection{Programmability of Hyper-Parameters (PoHP)}

Hyperparameters refer to the number and/or dimensionality of parameters of a feature such as the number of layers and size of filters in a CNN, or the number of histogram bins in HOG. They are usually determined at design time by using heuristics or through experiments. 

\smallskip\noindent\textbf{Advantages}: For learned features, changes in hyperparameters with proper (re-)training can result in significant performance improvements. As a result, hardware that supports PoHP can easily trade-off computational complexity for higher accuracy. For instance, a CNN object detection system with PoHP can choose between lower-complexity, lower-accuracy, e.g., AlexNet, and higher-complexity, higher-accuracy, e.g., VGG-16, according to the use cases. For hand-crafted features, however, the impact of PoHP on performance is usually limited, since it can be hard to capture the changes in parameter dimensionality without redesign. As a result, PoHP are not commonly supported for the hardware implementation of hand-crafted features.

\smallskip\noindent\textbf{Energy Cost}: Although PoHP can greatly benefits the performance of learned features, it comes at the price of lowered energy efficiency, since the hardware implementation must accommodates a wide range of possible configurations. First, it introduces \emph{irregular data movement}. For example, a CNN processor that supports PoHP has to deal with filters of many shapes and sizes, which complicates the control logic, synchronization schemes between parallel blocks, and data tiling optimization. Second, it usually requires \emph{higher dynamic range and resolution for data representation}, which have a negative impact on both computation and data movement. The bitwidth of datapath and the bandwidth requirement of memory all need to be designed for the worst-case scenario, which penalize the average energy efficiency.

\subsection{Programmability of Parameters (PoP)}

Parameters are the actual coefficients, such as the filter weights, that are used in the computation. For learned features, parameters are learned through training; for hand-crafted features, they are usually carefully designed according to the data statistics or pre-determined based on the desired properties. For instance, in HOG, the filter mask for the gradient extraction is simply $[-1~0~1]$.

\smallskip\noindent\textbf{Advantages}: With PoP, learned features can adapt to new data by simply retraining the parameters. For example, AlexNet has been shown to work for both RGB images and depth images~\cite{guptaECCV14}. For hand-crafted features, however, there is no PoP since all parameters are fixed at design time.

\smallskip\noindent\textbf{Energy Cost}: PoP also negatively impacts the hardware energy efficiency since the parameters have to be treated as data instead of fixed values during hardware implementation. This not only increases the required memory capacity and data movement, but also complicates the datapath design. In the case of CNN, the amount of parameters is usually too large to fit on-chip, which increases the accesses to energy-consuming off-chip memory, such as DRAM. In contrast, hand-crafted features can be greatly optimized for energy efficiency by hard-wiring the fixed parameters in datapaths or ROM. In the case of HOG, multiplications between the input images and gradient filter can be completely avoided. 


\section{Closing the Energy Gap}
\label{sec:closing_the_gap}

In Section~\ref{sec:hardware_implementation}, we have shown the energy efficiency results of hardware implementations from two extremes: the HOG hand-crafted feature with no PoP or PoHP, and the CNN learned feature with both PoP and PoHP, and the latter consumes an order of magnitude higher normalized energy per pixel than the former. A simple approach to reduce this energy gap is to remove all programmability in the hardware implementation of CNN, but this is not straightforward.

For example, the CONV layer weights from AlexNet can either be hard-wired in the multipliers, or stored in on-chip SRAM or ROM. This is not feasible if the available hardware resources are constrained to the level of the HOG design~\cite{suleiman201658}, i.e., 1000 kgates with 150 kB SRAM. Assuming each input and weight value take 1 byte, only 10k multipliers with fixed weights can be implemented in 1000 kgates, and only 150k weight values can fit in the SRAM. This number of multipliers cannot even fit 1\% of AlexNet 2334k weights in the CONV layers, and 15$\times$ larger memory is required to store all weights.

In this section, we will discuss some techniques that can be applied to reduce the energy efficiency gap.

\smallskip\noindent\textbf{Reduced Precision}:
Reducing data precision is an active research area for DNN~\cite{gysel2016hardware}, which directly reduces the worst-case bitwidth requirement when supporting PoHP. Specifically, 8-bit integer precision has become popular in recent DNN hardware~\cite{tpu}. Non-uniform quantization~\cite{han2016eie} and bitwidth reduction to 1-bit~\cite{nips2015-courbariaux-binaryconnect, courbariaux2016binarynet,eccv2016-rastegari-xnor_net} have also been demonstrated. Energy efficiency can also be improved if the hardware can adapt to the need of actual data. Custom datapath designs that adapt to the lower data precision show 2.56$\times$ energy savings in~\cite{moons20160}.


\smallskip\noindent\textbf{Sparsity}:
Increasing data sparsity reduces the intrinsic amount of computation and data movement, which improve energy efficiency. For CNN, the number of weights can be greatly reduced without reducing accuracy through pruning~\cite{nips1990-lecun-opt_brain_damage, nips2015-han, yang2016}. Specifically,~\cite{yang2016} has shown that the number of CONV layer weights in AlexNet can be reduced from 2334k to 352k, which is only double the memory capacity used in HOG; furthermore it reduces energy by 3.7$\times$. Specialized hardware designs can be used to exploit sparsity for increased speed or reduced energy consumption~\cite{isscc2016-chen,albericio2016cnvlutin, han2016eie}.


\smallskip\noindent\textbf{Data Compression}:
Sparsity also suggests opportunities for compression, which saves memory space and data transfer bandwidth. Many lightweight lossless compression schemes, such as run-length coding~\cite{isscc2016-chen} and Huffman coding~\cite{han2016eie}, are proposed to reduce the amount of off-chip data~\cite{isscc2016-chen,moons20160,han2016eie}. 


\smallskip\noindent\textbf{Energy Optimized Dataflow}:
PoHP incurs irregular data movement, preventing memory access optimization during hardware design. Therefore, designing hardware architecture that can adapt to the irregular data movement becomes critical to high energy efficiency, since data movement often consumes more energy than computation. Eyeriss demonstrates a reconfigurable architecture that can optimize data movement for various CNNs with a row stationary dataflow, and achieves 1.4$\times$ to 2.5$\times$ higher energy efficiency than existing designs~\cite{isca2016-chen}. 



Table~\ref{tab:techniques} summarizes the energy and memory savings using the discussed techniques. Combining them all can potentially deliver an order of magnitude reduction. Taking into account the fundamental computation gap discussed in Section~\ref{sec:hardware_implementation}, this reduction has the potential of closing the gap between HOG and CNN features.

\begin{table}
\centering
\small
\caption{CNN energy and memory savings using different techniques. *Measured on AlexNet **Assuming a 16-bit baseline}
\begin{tabular}{|c|c|c|} \hline
\textbf{Method}&\textbf{Energy}&\textbf{Memory Size}\\\hline
Reduced precision~\cite{tpu,moons20160}&2.56$\times$&2.0$\times$\\ \hline
Sparsity by pruning~\cite{yang2016}&3.7$\times$&6.6$\times$*\\ \hline
Data Compression~\cite{moons20160}&-&$>$2.0$\times$**\\ \hline
Energy optimized dataflow~\cite{isca2016-chen}&1.4--2.5$\times$&-\\ \hline
\end{tabular}
\label{tab:techniques}
\end{table}

\section{Conclusion}
\label{sec:conclusion}


The CNN learned features outperform the HOG hand-crafted features in visual object classification and detection tasks. This paper compares two chip designs for CNN and HOG to better understand the energy discrepancy between the two approaches and provide insight about the potential optimizations. Although learned features achieve more than 2$\times$ accuracy, it comes at a large 311$\times$ to 13,486$\times$ overhead in energy consumption. While a fundamental computation overhead exists, another order of magnitude gap is mainly caused by the fact that CNN architecture is programmable. A simple approach of removing all programmability in CNN and hard-wiring all multiplications doesn't work due to significant area cost (i.e., logic gates and on-chip memory).  Combing the techniques highlighted in the paper can potentially reduce the energy and memory sizes by an order of magnitude, and help reduce the gap between learned and hand-crafted features.


%


\bibliographystyle{ieeetr}
\scriptsize
\bibliography{__references}


\end{document}